\documentclass[10pt,twocolumn,letterpaper]{article}

\usepackage[pagenumbers]{wacv}

\definecolor{wacvblue}{rgb}{0.21,0.49,0.74}
\usepackage[pagebackref,breaklinks,colorlinks,allcolors=wacvblue]{hyperref}

\def\confName{}
\def\confYear{}

\title{AngioDG: Interpretable Channel-informed Feature-modulated Single-source Domain Generalization for Coronary Vessel Segmentation in X-ray Angiography}

\author{
Mohammad Atwany$^{1}$ \quad
Mojtaba Lashgari$^{1}$ \quad
Robin P.~Choudhury$^{2}$ \quad
Vicente Grau$^{1}$ \quad
Abhirup Banerjee$^{1,2}$\\[4pt]
$^{1}$Institute of Biomedical Engineering, Department of Engineering Science, University of Oxford, UK\\
$^{2}$Division of Cardiovascular Medicine, Radcliffe Department of Medicine, University of Oxford, UK\\[4pt]
{\tt\small \{mohammad.atwany@eng.ox.ac.uk,abhirup.banerjee@eng.ox.ac.uk\}}
}

\begin{document}
\maketitle
\begin{abstract}
Cardiovascular diseases are the leading cause of death globally, with X-ray Coronary Angiography (XCA) as the gold standard during real-time cardiac interventions. Segmentation of coronary vessels from XCA can facilitate downstream quantitative assessments, such as measurement of the stenosis severity and enhancing clinical decision-making. However, developing generalizable vessel segmentation models for XCA is challenging due to variations in imaging protocols and patient demographics that cause domain shifts. These limitations are exacerbated by the lack of annotated datasets, making Single-source Domain Generalization (SDG) a necessary solution for achieving generalization. Existing SDG methods are largely augmentation-based, which may not guarantee the mitigation of overfitting to augmented or synthetic domains. We propose a novel approach, ``AngioDG", to bridge this gap by channel regularization strategy to promote generalization. Our method identifies the contributions of early feature channels to task-specific metrics for DG, facilitating interpretability, and then reweights channels to calibrate and amplify domain-invariant features while attenuating domain-specific ones. We evaluate AngioDG on 6 x-ray angiography datasets for coronary vessels segmentation, achieving the best out-of-distribution performance among the compared methods, while maintaining consistent in-domain test performance.
\end{abstract}
    

\section{Introduction}\label{sec:intro}
Cardiovascular diseases were responsible for 20.5 million deaths globally in 2021 \cite{di2024heart}. X-ray Coronary Angiography (XCA)  is the current gold standard to assess and detect coronary stenosis (narrowing of coronary arteries supplying blood to the heart) during real-time cardiac interventions. XCA provides high-resolution images of the coronary arteries, which can be used for visual assessment and interventions such as balloon angioplasty or stent implantation.

The automatic segmentation of coronary vessels from x-ray angiograms presents significant potential to facilitate efficient and safe interpretations. It provides the foundation for several downstream tasks, such as 3D reconstruction of vascular trees \cite{banerjee2019point,wang2024deeplearningbased3dcoronary,ywang2024bioeng} enabling a more comprehensive assessment of stenosis and non-invasive calculation of virtual hemodynamic metrics to assess stenosis severity \cite{lashgari11patient}. The automatic segmentation of coronary vessels in XCA images is nevertheless challenging as it involves projecting a three-dimensional (3D) moving object onto a two-dimensional (2D) plane, producing imaging artifacts such as overlapping organs, unpredictable motion patterns, etc. 
Deep Learning (DL) models, particularly convolutional neural networks, have significantly outperformed classical approaches, though DL models often fail in real-world clinical settings due to the domain shift problem. Most DL models are trained and tested on data from the same distribution, assuming independent and identically distributed (i.i.d.) conditions \cite{he2022semistudentteacher,He2023Stacom}. However, deployment across different hospitals/equipment violates this assumption, leading to degradation on out-of-distribution (OOD) datasets. 

To address this domain shift, Domain Generalization (DG) has emerged as an effective approach, which assumes no access to the OOD test/target domain data. DG approaches can be divided into two main categories: Multi-source DG (MDG) that uses multiple source domains and Single-source DG (SDG), which relies on a single domain during training \cite{dg_survey}.
MDG approaches, though effective, often suffer from data collection challenges and privacy concerns, making SDG a more practical alternative \cite{yoon2023domainsurveymedicalimageanalysis}. However, SDG methods typically rely on data or feature augmentation, which may not guarantee robust generalization or prevent overfitting \cite{hu2023devilchannels}. Prior work demonstrates that first-layer feature channels are highly influential for SDG performance, as they encode the lowest-level features, which are often prone to capturing background noise, artifacts, and other domain-specific features \cite{hu2023devilchannels}. Inspired by this, we propose \textbf{AngioDG}, which combines channel decorrelation with channel reweighting to modulate early feature channels for SDG.

\subsection{Related Works on DG in Medical Image Segmentation}
DG has become crucial in medical image segmentation to address the domain shift between training and unseen test distributions. This is essential for deployment in diverse clinical settings, where varying equipment and acquisition protocols can lead to significant domain shifts, hindering model performance. 

DG methods in medical image segmentation can be broadly categorized into two main classes: MDG and SDG. MDG methods involve training the model on data from multiple source domains to improve model robustness across various domain distributions. They require access to multiple labeled data sources, which can be impractical in many medical settings. SDG methods, however, typically achieve generalization by expanding either the input or feature space in the shallow model layers, which does not guarantee mitigation of overfitting to domain-specific low-level features \cite{zhou2021domainmixstyle,hu2023devilchannels}. Examples of SDG methods for medical image segmentation include BigAug \cite{stackedtransformationtmi2020} that uses a series of stacked transformations to simulate domain shifts. Dual-Norm \cite{dualnormalizationcvpr2022} is another SDG method that augments source domain images into ``source-similar" images with similar intensities and ``source-dissimilar" images with intensity-inverted variants, and trains with different batch normalization (BN) layers to accommodate the two styles. Recent advancements also explored frequency-based domain augmentations to improve SDG \cite{frequencymixedmiccai2023}.

\begin{figure*}[!t]\centering
\includegraphics[width=0.85\textwidth]{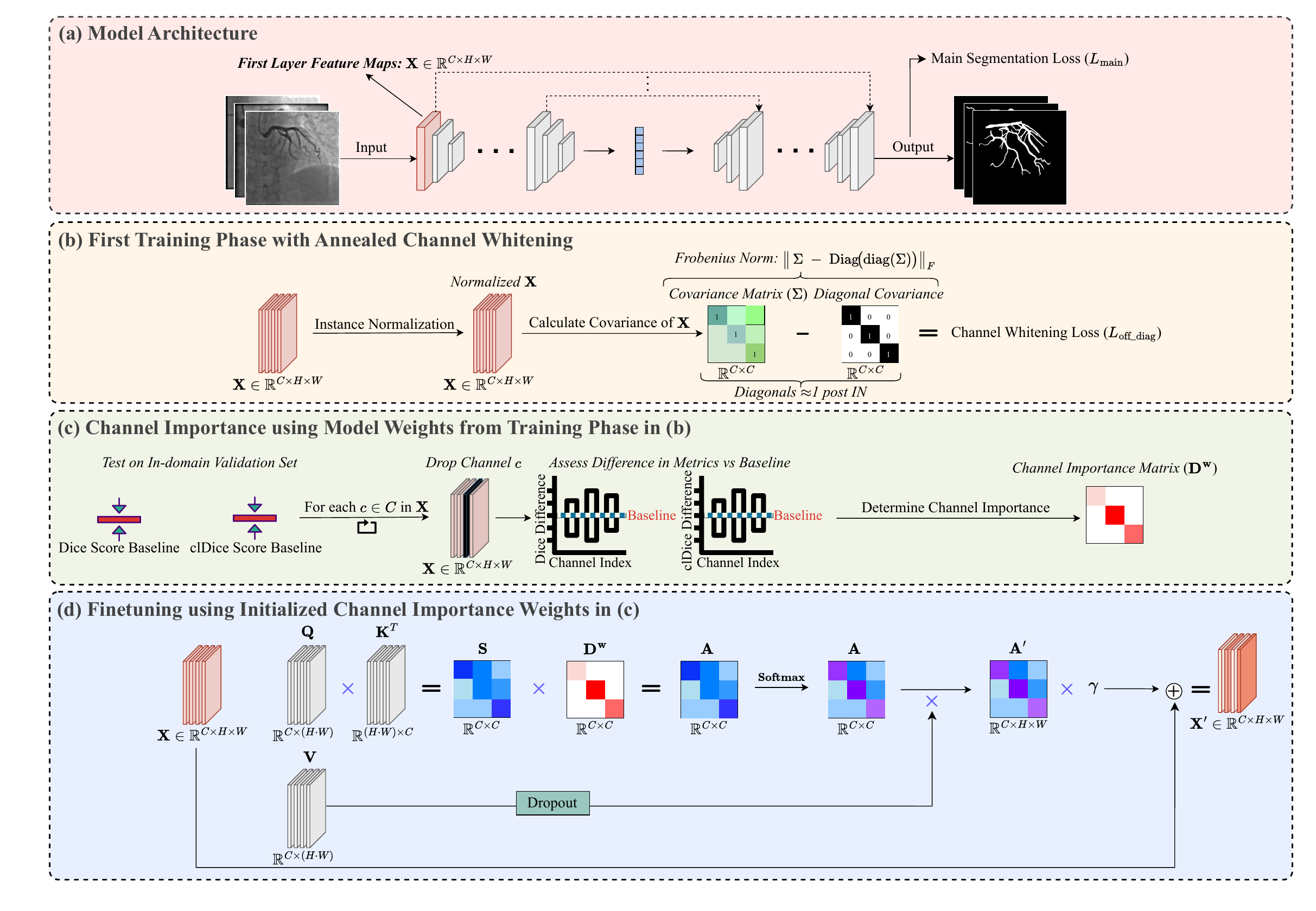}
\caption{Proposed AngioDG for SDG in coronary vessels segmentation from x-ray angiograms. (a) We use a U-Net with a ResNet-50 encoder and a 1024-channel decoder bottleneck (U-Net-R50-1k) as the model architecture. Feature maps (pre-BN) from the first convolutional layer (denoted as $\textbf{X}$ in (a)) are used in (b)-(d). (b) For initial training, in addition to the main segmentation loss, an annealed channel whitening loss is used to promote decorrelation of feature channels in the first layer. (c) Using pre-trained model weights from (b), we obtain channel importance by dropping channels and measuring the impact on Dice and clDice. Channels are weighted based on how their removal changes the Dice and clDice compared to the baseline scores obtained on the validation set. (d) In the final fine-tuning phase, we include WCA module after the first layer (pre-BN) using the channel-importance matrix $\mathbf{D}^{\mathbf{w}}$ (initialized from (c) and fine-tuned) to modulate contributions, in order to improve DG by emphasizing domain-invariant channels while suppressing domain-specific ones.}
\label{methodologyfigure} 
\end{figure*}

Despite these advancements, SDG methods are largely augmentation-based, aiming to cover a wider range of features to possibly cover more target domains. However, augmentation expands the data or feature space but does not directly assess whether the model overfits to such synthetic variations. A recent approach \cite{hu2023devilchannels} examined the contribution of individual shallow feature channels to SDG in optic disc/cup segmentation from fundus images, focusing on the first convolutional layer. This observation highlights the need to directly assess the influence of individual feature channels on SDG, as some channels may enhance generalization while others could contribute to overfitting.

\subsection{Related Works on Feature Channel Manipulation}

Building on recent insights in SDG literature, it is essential to  explore the common methods used for channel manipulation. Two relevant approaches in the literature are self-attention and Channel Whitening (CW).

Self-attention has proven to be a powerful tool in literature for capturing long-range dependencies and contextual information between features. Vaswani \textit{et~al.}\cite{NIPS2017_3f5ee243} initially introduced self-attention in the Transformer architecture for natural language processing applications, but its versatility soon led to its application across many DL domains. Fu \textit{et~al.} \cite{fu2019dualattentionpamcam} extended this concept by simultaneously applying self-attention to spatial positions and channels, effectively capturing spatial and channel-specific dependencies in images. This flexibility of self-attention makes it suitable for channel-wise operations in our method.

However, applying attention directly to feature channels can reinforce the existing inter-channel correlation structure, which may encode domain-specific style features and inadvertently amplify them. Hence, CW can serve as a precursor to channel attention by decorrelating channels, reducing reliance or the chances of amplification of domain-specific correlations. This is based on the assumption that domain-specific style often manifests in inter-channel correlations. A seminal example is instance-selective whitening \cite{Choi2021RobustNetwhitening}, which selectively whitens channels whose covariances are sensitive to photometric perturbations and hence deemed domain-specific, while preserving domain-invariant ones.

\subsection{Contributions}
In this work, we introduce ``AngioDG" to promote domain-invariant channels and mitigate domain-specific channels that hinder generalization for SDG in coronary vessels segmentation in x-ray angiography. AngioDG offers a targeted regularization approach to emphasize channel contributions in SDG. 
It uses a Weighted Channel Attention (WCA) \cite{fu2019dualattentionpamcam} module, which provides a direct approach to SDG  by assessing channel contributions to task-specific performance. Our contributions are summarized as follows:
\begin{itemize}
    \item The first part of AngioDG involves using CW in the first convolutional layer (pre-BN) to decorrelate channels. This step pushes the channels to learn decorrelated features during initial training.
    
    \item Next, AngioDG evaluates the contribution of each channel in the first layer (pre-BN) through channel-dropping to obtain channel importance weights, improving interpretability by identifying key features relevant to generalizable segmentation behavior. These weights are utilized with a modulating WCA module during fine-tuning to amplify domain-invariant channels and suppress domain-specific ones, thereby mitigating overfitting and improving generalization.
    
    \item By evaluating on 6 x-ray angiography datasets, AngioDG demonstrates improved OOD performance while maintaining consistent in-domain performance.

\end{itemize}

\section{Methodology}
\label{methodssection}
Our proposed AngioDG addresses the challenges in SDG for medical image segmentation by leveraging channel contributions, particularly in the first convolutional layer (pre-BN)  \cite{hu2023devilchannels}. This section describes the AngioDG methodology, implemented using a U-Net with a ResNet-50 encoder and a 1024-channel decoder bottleneck (U-Net-R50-1k), as shown in Fig.~\ref{methodologyfigure}(a). Here, we discuss the dataset/s, evaluation metrics, initial training using CW, channel importance identification, integration of the proposed WCA module for channel weighting/modulation, loss functions, and implementation details.

\subsection{Datasets}
We use six XCA image datasets, one as a source for training and five datasets as targets for OOD testing. Our source dataset is the SJTU dataset \cite{hao2020sequentialchinesedataset} with XCA images split into 173 for training, 86 for validation, and 69 for  testing from the Renji Hospital of Shanghai Jiao Tong University (SJTU) in China, with one labeled frame per subject with the most vessel visibility. For OOD testing, we test on five datasets: 1) the OxAMI (private) dataset collected from our local hospital containing 56 images annotated for the end-diastolic frame; 2) the UMAE T1-León (UMAE for short) dataset \cite{cervantes2019automaticmexicandataset} with 134 XCA images from Mexico, with annotations for the frame with most vessel visibility; 3) the CHUAC dataset~\cite{carballal2018automatic} containing 30 XCA images from the Complexo Hospitalario Universitario de A Coruña (CHUAC) in Spain; 4) the XCAD dataset \cite{ma2021selfXCADdataset} containing a total of 126 XCA images annotated by experts; and 5) the MTU dataset \cite{ZHAO2021104667314imagesdataset} with 616 XCA frames from the Jiangsu Province People’s Hospital, China, with manual annotations by cardiologists. We assess AngioDG's generalization on these datasets as they represent different patient demographics, imaging protocols, etc.

\subsection{Evaluation Metrics}
\label{evaluationmetrics}
We use two metrics to assess vessel segmentation performance: Dice score and centreline Dice (clDice) score \cite{shit2021cldice}. The Dice score measures the overlap between our predictions and ground truth for the vessel area. To evaluate connectivity preservation, we use clDice, which is the harmonic mean of topological sensitivity and topological precision.

\subsection{Initial Training and Channel Whitening}
\label{whiteninglossmethod}
\subsubsection{Instance-wise Channel Decorrelation}
We first train the model on the source domain and select the checkpoint that yields the maximum validation Dice as our baseline. We optimize for the main segmentation loss with a linearly annealed CW loss. The CW loss is applied to the raw first layer features (pre-BN) to push channels towards learning decorrelated features during training. We apply instance-wise channel whitening \cite{ulyanov2016instanceINstylization,Choi2021RobustNetwhitening} by minimizing the Frobenius-norm of the off-diagonal entries of each sample's channel covariance matrix \cite{cogswell2015reducingtotaldecorrelation,sun2016deepcoralaugust2025}. This pushes the covariance towards an identity-like matrix to minimize inter-channel correlations during training.

CW loss is calculated for the output of the first convolutional layer $\textbf{X}\in\mathbb{R}^{C \times H \times W}$ (pre-BN), where $C$ is the number of channels, $H$ is the height, and $W$ is the width. To calculate the CW loss, as in \cite{Choi2021RobustNetwhitening}, we apply Instance Normalization (IN) with no affine parameters to normalize each channel's mean to 0 and variance to $\approx$1 (IN computes per-sample per-channel statistics and uses a small $\varepsilon$ for numerical stability, so the empirical variance is slightly below 1).

After normalization, the output $\textbf{X}_{\text{norm}}$, which is already centered by the IN step, is reshaped (or flattened) into $\textbf{X}_{\text{flat}} \in \mathbb{R}^{C \times (H \cdot W)}$ to compute the covariance matrix $\Sigma$ as:
\begin{equation}
\Sigma = \frac{1}{(H \times W - 1)} \textbf{X}_{\text{flat}} \textbf{X}_{\text{flat}}^T.
\label{sigmacalculation}
\end{equation}

To decorrelate the channels, we remove the diagonal part of the covariance by subtracting it from $\Sigma$ and penalize only the off–diagonals, with a small $\varepsilon \mathbf{I}$ added to the diagonal beforehand for numerical stability:
\begin{equation}
\tilde{\Sigma} \;=\; \Sigma \;-\; \mathrm{Diag}\!\big(\mathrm{diag}(\Sigma)\big),
\qquad
L_{\text{off-diag}} \;=\; \big\| \tilde{\Sigma} \big\|_{F},
\label{offdiagonalloss}
\end{equation}
where $\mathrm{diag}(\Sigma)$ denotes the diagonal entries vector of $\Sigma$, and $\mathrm{Diag}(\cdot)$ maps a vector to a diagonal matrix with that vector on its diagonal. We compute this per instance and average over the batch.

\subsubsection{CW Loss Anneal}
\label{cwanneal}
During the first $e_0$ epochs (experimentally set at $e_0{=}15$), we disable the CW loss term $L_{\text{off-diag}}$ by setting its weight to $\beta=0$, while the main segmentation loss $L_{\text{main}}$ is optimized for the entire training epochs. Starting at epoch $e_0{+}1$, the CW loss is turned on at full weight $\beta=1$ and then linearly annealed to $0$ by the end of training. The total loss is
\begin{equation}
L_{\text{total}} \;=\; L_{\text{main}} \;+\; \beta(e)\,L_{\text{off-diag}},
\end{equation}
where $e$  is the epoch counter and $E$ is the total number of epochs in training. The annealing weight $\beta(e)$ follows a linear schedule:
\begin{equation}
\beta(e) =
\begin{cases}
0, & e \le e_0, \\[6pt]
1 - \dfrac{e - (e_0{+}1)}{E - (e_0{+}1)}, & e > e_0~.
\end{cases}
\end{equation}
This schedule allows stable early feature learning under the main segmentation loss before CW is applied, and later gradually reduces the influence of CW to avoid trading off feature decorrelation at the expense of segmentation performance.

\subsection{Channel Importance Estimation} 
\label{channelimportance}
\begin{figure}[!t]
\centering
\includegraphics[width=0.4\textwidth]{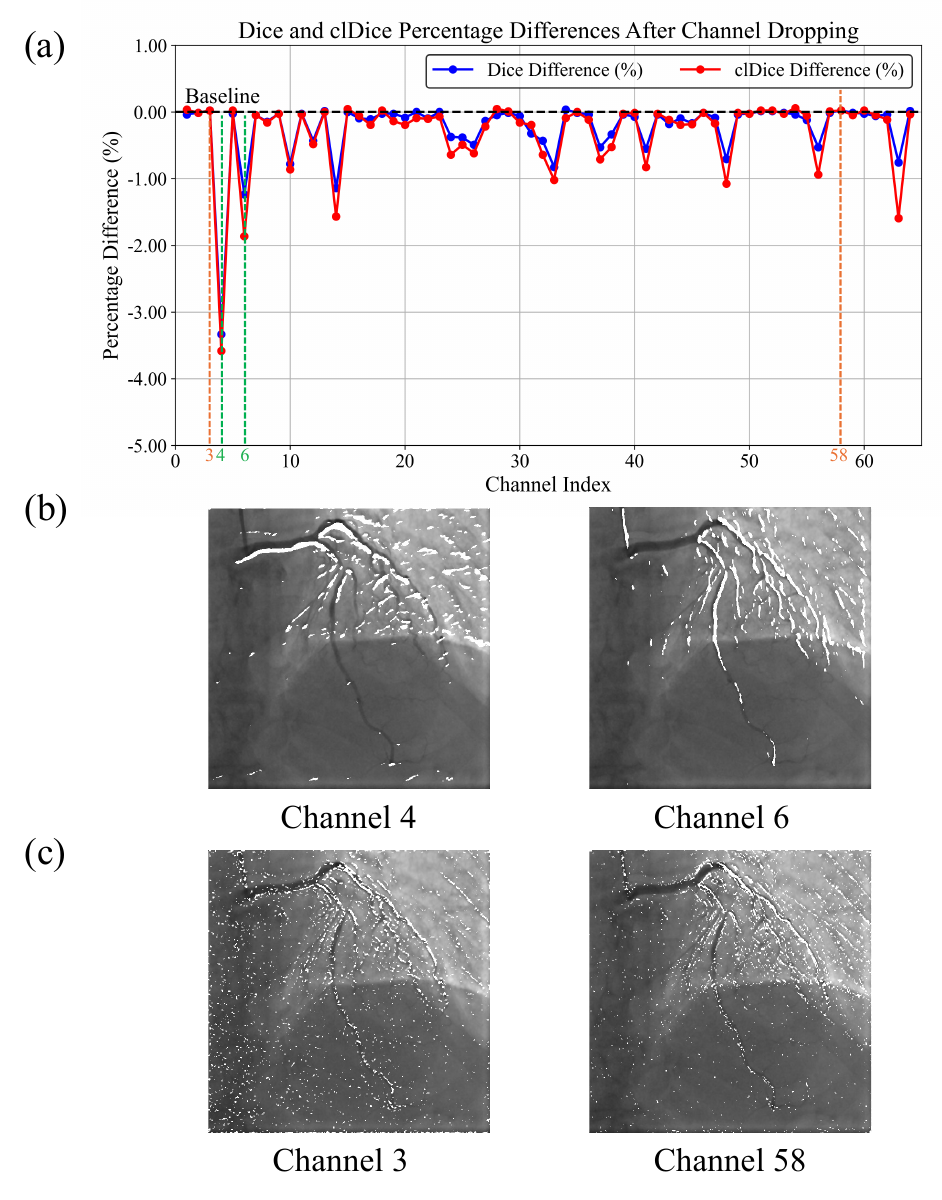}
\caption{Visualization of channel importance. We visualize each channel by overlaying the top $5\%$ of each channel’s activations ($\geq 95$th percentile) in white in (b) and (c):
(a) relative percentage change in Dice (Eq.~\ref{eq:delta_dice}) and clDice (Eq.~\ref{eq:delta_cldice}) scores on the internal validation set after dropping each channel;
(b) domain-invariant channels that capture vessel structures, shown in green dashed lines in (a);
(c) domain-specific channels that often capture noise or shadows, shown in orange dashed lines in (a).
}
\label{fig:channel_visualization}
\end{figure}

After initial training, we measure the importance of each channel in the first convolutional layer (pre-BN) as illustrated in Fig.~\ref{methodologyfigure}(c). We first load the pre-trained model weights from the initial training phase and evaluate the impact of dropping each channel at a time in the first layer on the validation set, using Dice and clDice as metrics. Figure~\ref{fig:channel_visualization}(a) shows the relative percentage change in Dice and clDice after dropping each individual channel, relative to the baseline with all channels active.

A channel is classified as domain-invariant if its removal results in performance deterioration compared to the baseline, indicating its importance for generalization across domains. As shown in Fig.~\ref{fig:channel_visualization}(b), these domain-invariant channels primarily capture the structural features of coronary vessels. Conversely, a channel is classified as domain-specific if its removal improves performance relative to the baseline, suggesting that the channel may have overfitted to domain-specific artifacts or noise primarily relevant to the training set. Figure~\ref{fig:channel_visualization}(c) illustrates how domain-specific channels often capture background noise or shadows.
Finally, a channel is considered neutral or redundant if its removal causes no change in performance, indicating potential redundancy with other channels in the feature map. Overall, in our SDG setting, the source dataset is the only observed domain during training; hence, we use the source validation-set channel-drop behavior as a heuristic, task-level indicator of generalizable domain-invariant features versus overfitted domain-specific features. 

To quantify each channel's importance, we compute the relative change in Dice and clDice scores for the validation set pre- and post-channel drop. The metric $\Delta D(\textbf{X}, i)$ is formulated as a weighted combination of the relative changes in Dice and clDice scores, as:
\begin{equation}
\Delta D(\mathbf{X}, i)
= \zeta\,\Delta Dice(\mathbf{X}, i)
+ (1-\zeta)\,\Delta clDice(\mathbf{X}, i),
\label{deltad}
\end{equation}
where $\zeta\in[0,1]$ trades off the Dice and clDice terms. Since Dice and clDice showcase similar trends on the validation set (Fig.~\ref{fig:channel_visualization}(a)), we set $\zeta=1$ for all experiments.

Here, $\Delta Dice(\textbf{X}, i)$ and $\Delta clDice(\textbf{X}, i)$ are defined as the relative changes in the Dice and clDice scores, respectively:
\begin{equation} 
\Delta Dice(\textbf{X}, i) = \frac{Dice(\textbf{X}^{i})-Dice(\textbf{X})}{Dice(\textbf{X})},
\label{eq:delta_dice}
\end{equation}
\begin{equation} 
\Delta clDice(\textbf{X}, i) = \frac{clDice(\textbf{X}^{i})-clDice(\textbf{X}) }{clDice(\textbf{X})},
\label{eq:delta_cldice}
\end{equation}
where $\textbf{X}^i$ is the feature map with the $i$-th channel of $\textbf{X}$, termed as $\textbf{X}_i$, set to zero.
The channel importance $\textbf{w}(\textbf{X}_i)$ is computed as:  
\begin{equation}  
\textbf{w}(\textbf{X}_i) =  
\begin{cases}  
1 + \log\left(1 + \left| \Delta D(\textbf{X}, i)\right| \right), &\text{if } \Delta D(\textbf{X}, i) < 0 \\  
1, &\text{if } \Delta D(\textbf{X}, i) = 0   \\
1 - \log\left(1 + \left| \Delta D(\textbf{X}, i) \right| \right), &\text{if } \Delta D(\textbf{X}, i) > 0.
\end{cases}  
\label{weightingchannels}  
\end{equation}

\subsection{Weighted Channel Attention (WCA)}
In our setting, we use CW as a preparatory step; we then use channel self-attention to modulate channels to improve SDG. After identifying channel importance in the first layer (pre-BN), we employ a WCA module that adapts the channel attention mechanism \cite{fu2019dualattentionpamcam} to enhance feature map contributions by weighting channels based on their importance in a second training phase (fine-tuning), as detailed in Fig.~\ref{methodologyfigure}(d).  The WCA module is only active during fine-tuning (BN layers are frozen) and disabled at test time.

The proposed WCA module first involves flattening the output of the raw first layer $\textbf{X}$ to $\mathbb{R}^{C \times (H \cdot W)}$ (pre-BN), which is used as query ($\textbf{Q}$), key ($\textbf{K}$), and value ($\textbf{V}$). We then calculate the similarity matrix $\textbf{S} = \textbf{Q} \times \textbf{K}^T$. The diagonal elements in $\textbf{S}$ represent the self-similarity of each channel, while the off-diagonal elements represent inter-channel similarities.    
Next, we obtain a weighted channel attention matrix $\textbf{A}$ by multiplying the \(\textbf{S}\) element-wise by a matrix \(\textbf{D}^\textbf{w} \in \mathbb{R}^{C \times C}\), where
\begin{equation}
{\textbf{D}^\textbf{w}}_{j,k} =
\begin{cases} 
\textbf{w}(\textbf{X}_i) & \text{if } j = k \\
1 & \text{if } j \neq k
\end{cases}
\end{equation}
We then subtract the maximum value in each row for numerical stability, scale by a factor ($\phi=0.10$ experimentally), and then apply softmax row-wise to normalize each row. We also apply dropout to the $\textbf{V}$ projection (\(p=0.20\)) using PyTorch  \texttt{nn.Dropout}. We then multiply $\textbf{A}$ with $\textbf{V}$ to emphasize the importance of features directly on the input to obtain $\textbf{A}'$. It is then scaled using a learnable parameter $\gamma$ (initialized at $\gamma=0.25$ experimentally) and added back to the input $\textbf{X}$ (residual connection). The learnable parameter $\gamma$ dynamically balances the effect of the attention-modified features and the original input. 

Both $\gamma$ and the diagonal weights $\textbf{w}(\textbf{X})$ (initialized as per Sec.~\ref{channelimportance}) are set as trainable after their initialization for fine-tuning. We also unfreeze a shallow subset (first pre-BN layer and next convolutional block) with a smaller learning rate in order to allow room for dynamic calibration of low-level filters while still being influenced by the prior induced by the WCA. Since we use the same source dataset for fine-tuning as initial training, we freeze  BN layers during fine-tuning to avoid the running statistics overfitting to the training domain.

\subsection{Losses}
\label{losses_section}
Our training involves two phases: initial training and fine-tuning. In initial training, we optimize the model for two losses: CW loss ($L_{\text{off-diag}}$) on  the first layer (pre-BN) channels annealed as in Sec.~\ref{cwanneal} and connectivity-based segmentation loss ($L_{\text{main}}$) (as per Yang~\textit{et al.}'s \cite{yang2023directionaldconnet} official code). In fine-tuning, we only include $L_{\text{main}}$.

\begin{table*}
\centering
\caption{Performance evaluation of the ablation study on AngioDG components, in terms of Dice score (\%), using the SJTU dataset as source domain. LA@$e_0$ refers to applying delayed linear anneal, defined in Sec.~\ref{cwanneal}, starting after epoch $e_0$. Best results of each dataset are annotated as  \textbf{bold}, and second best results as \textit{italics}.}
\label{finaldicescores_ablation}
\setlength{\tabcolsep}{4pt} 
\begin{tabular}{lllllll}
\hline
Model & SJTU (In-domain) & OxAMI & UMAE & CHUAC & XCAD & MTU \\
\hline
Baseline & 83.27 $\pm$ 2.94 & 78.87 $\pm$ 7.79 & \textit{69.12 $\pm$ 5.82} & 69.09 $\pm$ 6.50 & 75.34 $\pm$ 5.30 & 82.35 $\pm$ 6.71 \\
Baseline  + CW (LA@0) & 83.55 $\pm$ 2.78 & 78.66 $\pm$ 8.96 & 68.73 $\pm$ 6.77 & \textit{70.98 $\pm$ 5.05} & 75.37 $\pm$ 5.12 & \textit{82.75 $\pm$ 5.99} \\
Baseline + CW (LA@5) & 83.54 $\pm$ 3.05 & 77.66 $\pm$ 9.13 & 68.54 $\pm$ 7.90 & 69.44 $\pm$ 6.09 & 75.11 $\pm$ 4.49 & 82.05 $\pm$ 6.67 \\
Baseline  + CW (LA@10) & 83.68 $\pm$ 2.98 & 77.89 $\pm$ 8.94 & 68.16 $\pm$ 7.08 & 70.05 $\pm$ 5.54 & 75.34 $\pm$ 5.20 & 82.42 $\pm$ 6.77 \\
Baseline  + CW (LA@15) & \textit{83.78 $\pm$ 3.05} &\textit{78.98 $\pm$ 8.08} & 68.12 $\pm$ 7.07 & 70.87 $\pm$ 5.11 & \textit{75.71 $\pm$ 5.13} & 82.61 $\pm$ 6.54 \\
Baseline  + CW (LA@20) & 83.62 $\pm$ 3.18 & 78.72 $\pm$ 8.12 & \textbf{69.52 $\pm$ 5.90} & 69.99 $\pm$ 6.27 & 74.99 $\pm$ 5.25 & 82.30 $\pm$ 6.82 \\
AngioDG & \textbf{83.89 $\pm$ 2.97} & \textbf{79.38 $\pm$ 8.03} & 68.90 $\pm$ 6.79 & \textbf{71.55 $\pm$ 5.29} & \textbf{75.91 $\pm$ 5.09} & \textbf{82.76 $\pm$ 7.15} \\
\hline
\end{tabular}
\end{table*}

\begin{table*}
\centering
\caption{Performance evaluation of the ablation study on AngioDG components, in terms of clDice score (\%), using the SJTU dataset as source domain. LA@$e_0$ refers to applying delayed linear anneal, defined in Sec.~\ref{cwanneal}, starting after epoch $e_0$. Best results of each dataset are annotated as  \textbf{bold}, and second best results as \textit{italics}.}
\label{finalcldicescores_ablation}
\setlength{\tabcolsep}{4pt} 
\begin{tabular}{lllllll}
\hline
Model & SJTU (In-domain) & OxAMI & UMAE & CHUAC & XCAD & MTU \\
\hline
Baseline & 87.74 $\pm$ 3.09 & 82.53 $\pm$ 9.23 & \textbf{80.93 $\pm$ 6.28} & 77.09 $\pm$ 6.53 & 78.42 $\pm$ 6.48 & 84.11 $\pm$ 7.54 \\
Baseline + CW (LA@0)  & 88.04 $\pm$ 2.94 & 82.71 $\pm$ 10.28 & 79.85 $\pm$ 7.55 & 78.04 $\pm$ 5.81 & 78.57 $\pm$ 6.50 & 83.97 $\pm$ 7.12 \\
Baseline + CW (LA@5) & \textit{88.39 $\pm$ 2.96} & 81.26 $\pm$ 10.54 & 80.04 $\pm$ 8.87 & 77.03 $\pm$ 6.39 & 78.09 $\pm$ 5.94 & 83.62 $\pm$ 7.75 \\
Baseline + CW (LA@10) & 88.25 $\pm$ 2.92 & 81.77 $\pm$ 10.54 & 79.75 $\pm$ 7.96 & 78.10 $\pm$ 5.91 & \textit{78.97 $\pm$ 6.51} & \textit{84.12 $\pm$ 7.57} \\
Baseline + CW (LA@15) & \textbf{88.47 $\pm$ 3.04} & \textit{83.18 $\pm$ 9.26} & 80.43 $\pm$ 8.01 & \textit{78.15 $\pm$ 5.49} & \textit{78.97 $\pm$ 6.60} & 84.07 $\pm$ 7.42 \\
Baseline + CW (LA@20) & 88.13 $\pm$ 3.37 & 82.74 $\pm$ 9.38 & \textit{80.71 $\pm$ 6.40} & 77.37 $\pm$ 6.41 & 77.99 $\pm$ 6.69 & 83.77 $\pm$ 7.77 \\
AngioDG & 88.36 $\pm$ 3.04 & \textbf{83.41 $\pm$ 9.12} & 80.62 $\pm$ 7.66 & \textbf{78.78 $\pm$ 5.44} & \textbf{79.18 $\pm$ 6.41} & \textbf{84.22 $\pm$ 7.84} \\
\hline
\end{tabular}
\end{table*}

\subsection{Implementation Details}
We use the SJTU dataset as the source with its predefined train, validation, and test split, and evaluate on the in-domain test set and five OOD datasets. For all models we use an Adam optimizer and a batch size of 8. For initial training, we use a polynomial learning rate (LR) scheduler (set for total iterations but stepped once per epoch for small per-epoch decays) and train for 400 epochs with an initial LR of 1e-4; the first convolutional layer (pre-BN) uses  $10\times$ the base LR as it is optimized for both the main segmentation loss and the CW loss. All experiments, including comparisons with augmentation-based DG methods, use identical settings (including the $10\times$ base LR for first layer) for fair comparison, and we report the checkpoint with the best validation Dice.  All first layer operations in  AngioDG (CW, importance estimation, WCA) and integrated augmentation-based methods, where applicable, operate on the pre-BN ``\texttt{conv1}" features; investigating deeper placement or post-BN is saved for future work. We then fine-tune for 20 epochs (details in \textbf{Supplementary, Sec.~B}) and select the best checkpoint using a validation-Dice plateau-stability criterion.

\begin{table*}
\centering
\small
\setlength{\tabcolsep}{2.5pt}
\caption{Performance evaluation of augmentation-based SDG methods with AngioDG, in terms of Dice score (\%), using the SJTU dataset as source domain. Best results of each dataset are annotated as \textbf{bold}, and second best results as \textit{italics}.}
\label{finaldicescores_sdg_new}
\begin{tabular}{lllllllll}
\hline
Model & SJTU (In-domain) & OxAMI & UMAE & CHUAC & XCAD & MTU & Avg. OOD & Avg. All \\
\hline
MixStyle  & 83.94 $\pm$ 2.84 & 78.73 $\pm$ 7.54 & 68.48 $\pm$ 5.88 & 69.00 $\pm$ 5.29 & 74.96 $\pm$ 5.37 & 82.51 $\pm$ 5.98 & 74.74 $\pm$ 6.09 & 76.27 $\pm$ 6.62 \\
EFDMix    & 83.97 $\pm$ 2.68 & \textit{78.99 $\pm$ 7.34} &\textbf{69.11 $\pm$ 5.94} & 71.01 $\pm$ 5.32 & 75.20 $\pm$ 5.46 & \textbf{83.11 $\pm$ 5.36} & \textit{75.48 $\pm$ 5.73} & \textit{76.90 $\pm$ 6.18} \\
TriD      & \textbf{84.17 $\pm$ 2.77} & 76.01 $\pm$ 12.10 & 67.71 $\pm$ 5.99 & \textbf{71.98 $\pm$ 4.71} & 75.47 $\pm$ 5.49 & 81.49 $\pm$ 7.37 & 74.53 $\pm$ 5.11 & 76.14 $\pm$ 6.03 \\
FourierAM & \textit{84.15 $\pm$ 2.80} & 75.31 $\pm$ 14.53 & 68.47 $\pm$ 6.97 & 71.13 $\pm$ 6.00 & \textit{75.74 $\pm$ 5.47} & 82.37 $\pm$ 10.05 & 74.60 $\pm$ 5.29 & 76.20 $\pm$ 6.13 \\
InstMix   & 83.15 $\pm$ 3.45 & 77.79 $\pm$ 9.15 & 67.29 $\pm$ 7.46 & 70.95 $\pm$ 6.71 & 74.82 $\pm$ 5.12 & 81.91 $\pm$ 7.56 & 74.55 $\pm$ 5.71 & 75.99 $\pm$ 6.20 \\
AngioDG   & 83.89 $\pm$ 2.97 & \textbf{79.38 $\pm$ 8.03} & \textit{68.90 $\pm$ 6.79} & \textit{71.55 $\pm$ 5.29} & \textbf{75.91 $\pm$ 5.09} & \textit{82.76 $\pm$ 7.15} & \textbf{75.70 $\pm$ 5.63} & \textbf{77.07 $\pm$ 6.05} \\
\hline
\end{tabular}
\end{table*}

\begin{table*}
\centering
\small
\setlength{\tabcolsep}{2.5pt}
\caption{Performance evaluation of augmentation-based SDG methods with AngioDG, in terms of clDice score (\%), using the SJTU dataset as source domain. Best results of each dataset are annotated as \textbf{bold}, and second best results as \textit{italics}.}
\label{finalcldicescores_sdg_new}
\begin{tabular}{lllllllll}
\hline
Model & SJTU (In-domain) & OxAMI & UMAE & CHUAC & XCAD & MTU & Avg. OOD & Avg. All \\
\hline
MixStyle  & 88.44 $\pm$ 2.96 & 82.63 $\pm$ 8.99 & 79.76 $\pm$ 6.27 & 77.39 $\pm$ 5.46 & 77.41 $\pm$ 6.99 & 84.16 $\pm$ 7.10 & 80.27 $\pm$ 3.06 & 81.63 $\pm$ 4.31 \\
EFDMix    & 88.50 $\pm$ 2.76 & \textit{82.97 $\pm$ 8.60} & 80.22 $\pm$ 6.41 & \textit{78.72 $\pm$ 6.05} & 77.97 $\pm$ 6.68 & \textbf{84.92 $\pm$ 6.47} & \textit{80.96 $\pm$ 2.93} & \textit{82.22 $\pm$ 4.04} \\
TriD      & \textit{88.72 $\pm$ 3.03} & 79.90 $\pm$ 12.29 & 78.22 $\pm$ 7.48 & 78.10 $\pm$ 6.21 & \textbf{79.30 $\pm$ 6.66} & 82.66 $\pm$ 8.41 & 79.64 $\pm$ 1.85 & 81.15 $\pm$ 4.06 \\
FourierAM & \textbf{88.89 $\pm$ 2.79} & 77.60 $\pm$ 15.98 & \textit{80.50 $\pm$ 7.29} & 77.66 $\pm$ 6.42 & 78.65 $\pm$ 6.58 & 83.68 $\pm$ 10.54 & 79.62 $\pm$ 2.56 & 81.16 $\pm$ 4.42 \\
InstMix   & 87.54 $\pm$ 3.52 & 80.57 $\pm$ 10.10 & 78.99 $\pm$ 8.36 & 78.08 $\pm$ 6.82 & 76.87 $\pm$ 6.91 & 82.80 $\pm$ 8.93 & 79.46 $\pm$ 2.30 & 80.81 $\pm$ 3.89 \\
AngioDG   & 88.36 $\pm$ 3.04 & \textbf{83.41 $\pm$ 9.12} & \textbf{80.62 $\pm$ 7.66} & \textbf{78.78 $\pm$ 5.44} & \textit{79.18 $\pm$ 6.41} & \textit{84.22 $\pm$ 7.84} & \textbf{81.24 $\pm$ 2.46} & \textbf{82.43 $\pm$ 3.65} \\
\hline
\end{tabular}
\end{table*}

\section{Results}
\label{mainresultssection}
For evaluating the performance of AngioDG, we conduct: 1) ablation studies to examine the roles of CW and WCA module; and 2) comparison with diverse augmentation-based DG methods (reproduced by integrating them into our model in an SDG setting for fairness) each representing a unique strategy -- style mixing in MixStyle \cite{zhou2021domainmixstyle}, exact feature distribution matching in EFDMix \cite{zhang2022exactefd}, domain randomization in TriD \cite{treasureind}, Fourier Amplitude Mixing (FourierAM) following \cite{xu2021FACT}, and a control baseline for instance-level feature averaging, referred to as ``InstMix". For fairness, we use identical training settings for all compared methods (details in \textbf{Supplementary, Sec.~C.2}).

\subsection{Ablation Study}
As reported in Tables~\ref{finaldicescores_ablation} and \ref{finalcldicescores_ablation} for Dice and clDice, respectively, we ablate the CW delayed-linear-anneal start epoch, LA@$e_0$ (annealing starts after epoch $e_0$), and observe that $e_0=15$ yields the most balanced setting in terms of mean and standard deviation (SD), and hence use it for subsequent fine-tuning with WCA in AngioDG.

In terms of Dice score, for the in-domain test set (SJTU), LA@15 results in a relative improvement of  $+0.61\%$ over the baseline, with AngioDG nudging it further but with a more stable SD $-2.62\%$. On the OxAMI dataset, LA@15 slightly improves over the baseline, with AngioDG further improving LA@15 by $+0.51\%$ (SD $-0.62\%$). On the UMAE dataset, LA@15 slightly decreases the mean, but AngioDG recovers with an improvement of $+1.15\%$ and SD $-3.96\%$ relative to LA@15. Stronger gains are observed on the CHUAC dataset, where LA@15 yields a relative improvement of $+2.58\%$ with an SD reduction of $-21.38\%$ relative to baseline;  AngioDG then further improves the mean by $+0.96\%$. For the XCAD dataset, LA@15 yields an improvement of $+0.49\%$ (SD $-3.21\%$), and AngioDG improves by $+0.26\%$ (SD $-0.78\%$). For the MTU dataset, LA@15 improves by $+0.32\%$ (SD $-2.53\%$) over baseline and AngioDG improves by $+0.18\%$.

In terms of clDice, for the in-domain test set (SJTU), LA@15 results in a relative improvement of  $+0.83\%$ (SD $-1.62\%$) over baseline, while AngioDG yields a comparable mean with SD unchanged. On the OxAMI dataset, LA@15 improves by $+0.79\%$, then  AngioDG further improves it by  $+0.28\%$ (SD $-1.51\%$) relative to  LA@15. On the UMAE dataset, LA@15 slightly decreases the mean, but AngioDG improves it up by $+0.24\%$ (SD $-4.37\%$) relative to LA@15. The CHUAC dataset sees larger improvements: LA@15 improves by $+1.38\%$ (SD $-15.93\%$) over baseline and AngioDG further improves by $+0.81\%$ (SD $-0.91\%$). The XCAD dataset sees a relative improvement of $+0.70\%$ by LA@15 and then  $+0.27\%$ (SD $-2.88\%$) by AngioDG. The MTU dataset remains stable in mean and SD. Ultimately, in terms of ablation, AngioDG achieves the highest OOD Dice and clDice results on 4 out of 5 datasets.

\subsection{Comparison with Augmentation-based SDG Methods}
We compare AngioDG with MixStyle, EFDMix, TriD, FourierAM, and InstMix (simple baseline) as reported in Tables~\ref{finaldicescores_sdg_new} and \ref{finalcldicescores_sdg_new} for Dice and clDice, respectively. In our experiments,  AngioDG achieves the best OOD average for both Dice at $75.70$ and clDice at $81.24$. It also achieves the best overall average with a Dice score of $77.07$ and clDice score of $82.43$ (we report all scores in this subsection as percentages). 

\begin{figure*}[!t]\centering
\includegraphics[width=0.60\textwidth]{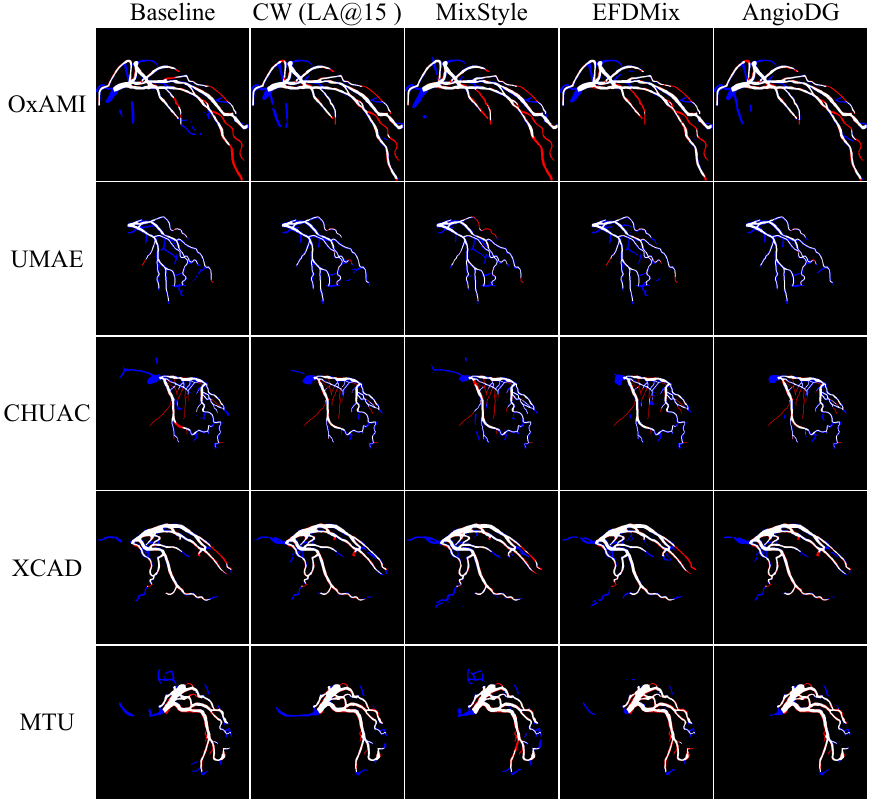} 
\caption{Qualitative performance analysis of coronary vessels segmentation across OOD  datasets for ablation components, AngioDG, MixStyle, and EFDMix. True positive, false positive, and false negative are shown in white, blue, and red, respectively. The incorporation of different components in AngioDG demonstrates improvements in vessel delineation and generalization.}
\label{qualitativefinalresults} 
\end{figure*}

On the in-domain test set, AngioDG achieves a competitive Dice score of $83.89$ and clDice score of $88.36$ compared to augmentation-based methods. In terms of OOD evaluation, AngioDG achieves the highest Dice and clDice on the OxAMI  dataset with $79.38$ and $83.41$, respectively. On the UMAE dataset, AngioDG achieves the second-highest Dice of $68.90$ and the highest clDice of $80.62$. On the CHUAC dataset, AngioDG achieves the second-highest Dice of $71.55$ and the highest clDice  of $78.78$. On the XCAD dataset, AngioDG achieves the highest Dice of $75.91$ and the second-highest clDice of $79.18$. For the MTU dataset, AngioDG achieves the second-highest    Dice and clDice of $82.76$ and $84.22$, respectively. Overall, AngioDG achieves the highest Dice on two and highest clDice on three out of five OOD datasets, reflecting strong generalization while maintaining in-domain performance. Qualitative results in Fig.~\ref{qualitativefinalresults} reflect these improvements, showing better vessel overlap and connectivity.

\section{Discussion}
\subsection{Ablation Study}
Our ablation study shows that delaying CW with linear decay generally improves generalization by decorrelating feature channels  after a short warm-up. As suggested by the results, LA@15 balances warm-up for stable early training and the regularization period, with linear decay maintaining focus on the main segmentation loss. Importantly, we employ CW to reduce the chances of WCA inadvertently amplifying domain-specific correlations.

In practice, this manifests as a more spread out channel importance; under the same fine-tuning setting, we observe that LA@15 improves performance, while fine-tuning the baseline (no CW) leads to suboptimal results. We hypothesize that this is due to a mix of feature scale effects and concentrated channel importances, as observed in channel-drop curves (\textbf{Fig.~S1 in Supplementary}), which show that the baseline over-relies on a few first layer channels, whereas LA@15 results in smaller and relatively more consistent drops.  While additional hyperparameter tuning in the fine-tuning phase  (e.g., softmax scaling $\phi$, etc.) may help, the observations suggest that CW spreads out channel importance and reduces over-reliance, which aligns with better OOD performance. These observations further motivate using LA@15 weights for subsequent  WCA fine-tuning.

After CW, fine-tuning the early layers (first raw layer and second block) with WCA calibrates previously learned per-channel contributions. During fine-tuning, we use the same train/validation split as initial training, so changes in validation Dice are subtle. Hence, we hypothesize that OOD improvements are attributed to CW, followed by channel emphasis combined with selecting a more stable solution (weights) in the early layers via a plateau-based checkpoint rule. In practice, we observe a very slightly lower validation Dice based on the selection criteria in fine-tuning but better OOD generalization, which suggests a generalization-stability trade-off.

\subsection{Comparison with Augmentation-based SDG Methods}
Our comparative analysis shows that AngioDG outperforms augmentation-based methods in terms of overall average and OOD average, reflecting both in-domain consistency and DG performance gains. As these approaches rely on expanding the feature space by introducing synthetic variations for DG, the introduced variations might not directly align with all target distributions and may explain the mixed OOD gains observed. Nevertheless, improved DG performance of AngioDG can be attributed to its ability to directly emphasize domain-invariant features in an interpretable way, while down-weighting domain-specific ones supported by a more stable solution during fine-tuning. Also, since WCA is employed in fine-tuning only (BN frozen) and not during test time, improved performance can be linked to a better modulation of low-level features (first raw layer and second block) rather than additional inference-time layers. Improvements in Dice score highlight the emphasis on preserving vessel width, whereas improvements in clDice reflect emphasis on connectivity.

Notably, while AngioDG does not perform DG-oriented augmentation in the input space or feature space, it instead relies on source-validation channel-drop behavior as a task-level proxy for Dice and clDice, yet it still slightly surpasses the augmentation-based EFDMix (that augments three shallow feature positions) for OOD and overall Dice and clDice in our experiments. This highlights the potential of targeting individual channel modulation as a  promising direction to explore alongside other SDG approaches.

\section{Conclusion}
In this work, we introduce AngioDG -- a novel SDG method for coronary vessels segmentation in x-ray angiography. AngioDG employs channel whitening followed by a novel WCA module to explicitly address feature channel learning through the interpretation of their feature contributions, mitigating overfitting and improving generalization. By amplifying domain-invariant channels and suppressing domain-specific ones, AngioDG attains the best averages among the compared methods on 5 OOD XCA datasets, while maintaining consistent in-domain performance. This generalizability enhances the robust application of the proposed AngioDG for real-time cardiac interventions, which can support non-invasive coronary disease severity quantification through 3D vessel reconstruction and \textit{in silico} studies \cite{lashgari11patient}.

{
    \small
    \bibliographystyle{ieeenat_fullname}
    \bibliography{refs}
}

\end{document}


\title{Supplementary Material for the paper titled AngioDG: Interpretable Channel-informed Feature-modulated Single-source Domain Generalization for Coronary Vessel Segmentation in X-ray Angiography}  
\author{Same authors as main paper}

\maketitle
\thispagestyle{empty}
\appendix

\section{Channel-drop Interpretation}
When estimating channel importance in the first layer, we drop pre-BN channels by setting them to 0, which yields spatially constant maps after BN (since BN uses running statistics during evaluation), which we consider as effectively removing spatial structure from that channel.  We therefore look at the changes in metrics such as Dice and clDice as a proxy for the dropped channel's contribution towards spatial content. We leave the exploration of post-BN channel dropping or even deeper layers for future work.

\section{WCA Fine-tuning}
During fine-tuning (for 20 epochs), we use a stable validation-Dice plateau criterion (last epoch of the first stable plateau with Dice within \( \epsilon \) of the running best), because fine-tuning calibrates channel weights on the same train/validation split, so improvements are small and we want to choose the most stable epoch. We also use a \texttt{ReduceLROnPlateau} LR scheduler driven by validation loss; we initialize parameter groups as: WCA parameters with 1e-3 and \texttt{conv1}+block2 parameters with 1e-4.

For validation checkpoint selection in fine-tuning,  we use the last epoch of the first stable validation plateau and save its weights, i.e., within a small adaptive tolerance $\epsilon$ of the running-best validation Dice while also exhibiting low variance in a short window (or low SD). We allow for a small warm-up period before applying the selection criterion. The adaptive $\epsilon$ is calculated from a short rolling window of recent validation Dice values using the median absolute deviation (MAD). We therefore save checkpoints while on the  first stable plateau and lock the last checkpoint when we effectively leave the plateau. 

Moreover, as we fix the fine-tuning epochs to 20 (5\% of the total 400), we observe that our fine-tuning of LA@15 is already on a plateau by epoch 20 (it might not have exited the plateau yet). Hence, as ablation, we extend training for 5 more epochs and observe marginal changes, indicating model convergence; we report these results in Table~\ref{tab:extended25epochs}. For the case of Baseline + WCA we see no difference in extending epochs (plateau epoch selection at epoch 11).

\begin{table*}[!htbp]
\centering
\small
\caption{Extended training results for the fine-tuning of AngioDG for 25 epochs.}
\label{tab:extended25epochs}
\resizebox{\textwidth}{!}{
\begin{tabular}{l cc cc cc cc cc cc}
\toprule
& \multicolumn{2}{c}{SJTU (In-domain)} 
& \multicolumn{2}{c}{OxAMI} 
& \multicolumn{2}{c}{UMAE} 
& \multicolumn{2}{c}{CHUAC} 
& \multicolumn{2}{c}{XCAD} 
& \multicolumn{2}{c}{MTU} \\
\cmidrule(lr){2-3} \cmidrule(lr){4-5} \cmidrule(lr){6-7} \cmidrule(lr){8-9} \cmidrule(lr){10-11} \cmidrule(lr){12-13}
Method 
& Dice & clDice & Dice & clDice & Dice & clDice & Dice & clDice & Dice & clDice & Dice & clDice \\
\midrule
AngioDG (25 epochs) 
& 83.85 $\pm$ 3.00 & 88.42 $\pm$ 3.09
& 79.47 $\pm$ 7.86 & 83.54 $\pm$ 9.00
& 68.77 $\pm$ 6.81 & 80.54 $\pm$ 7.69
& 71.42 $\pm$ 5.24 & 78.63 $\pm$ 5.42
& 75.83 $\pm$ 5.12 & 79.05 $\pm$ 6.46
& 82.81 $\pm$ 6.83 & 84.32 $\pm$ 7.50 \\
\bottomrule
\end{tabular}
}
\end{table*}

\section{Implementation Details}
\subsection{Data Pre-processing and Augmentations}
All input XCA images are  grayscale with spatial size $(512, 512)$, except for 1) the UMAE T1-León dataset images, which are initially $(300, 300)$ and hence we zero-pad to $(512, 512)$; 2) the CHUAC dataset, where images are of size $(189,189)$ and therefore bilinearly upsampled to size $(512,512)$ to match their ground truth masks. We perform a rebinarization check on all ground truth masks for strictness. We also apply augmentations during training, including: 1) color adjustments using hue $(-30^\circ, 30^\circ)$, saturation $(-5, 5)$, and value shift $(-15, 15)$; 2) geometric transformations including horizontal and vertical flipping, random \(90^{\circ}\) rotations, random shift $(-0.05, 0.05)$, scaling $(-0.1, 0.1)$, aspect ratio changes $(-0.1, 0.1)$, and rotation $(-30^{\circ}, 30^{\circ})$ (OpenCV bilinear resampling is applied in a \texttt{randomShiftScaleRotate} function only, with rebinarizing the masks at 0.5; this only affects training and not evaluation).

\begin{figure*}[!htbp]\centering
\includegraphics[width=\textwidth]{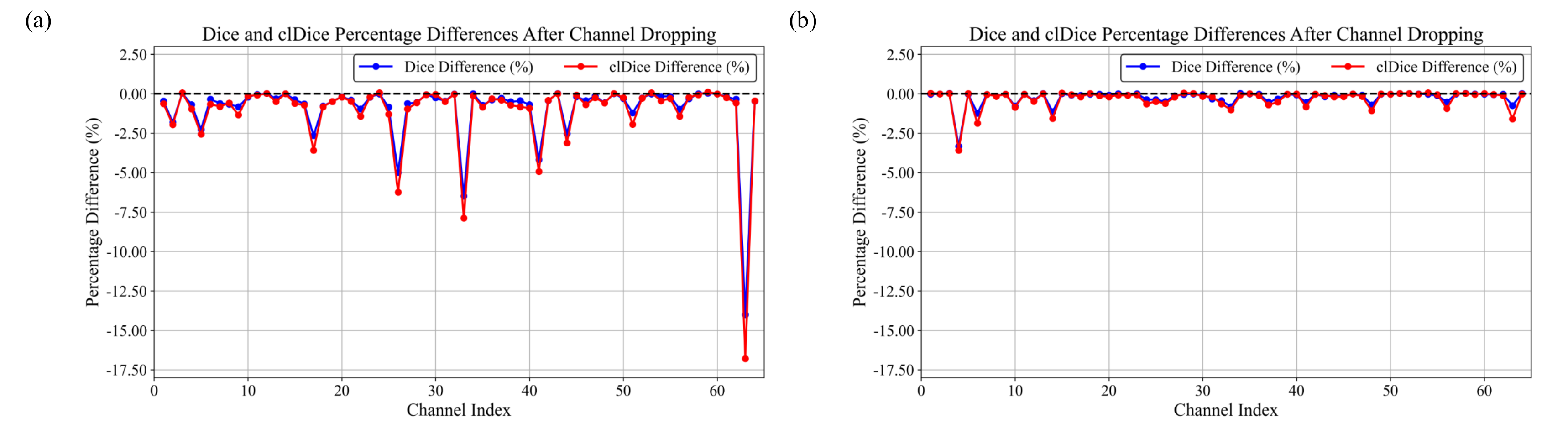}
\caption{Relative percentage differences in terms of Dice and clDice for dropping the first layer (pre-BN) channels for a) Baseline and b) LA@15 (CW with delayed linear annealing active after the 15th epoch).}
\label{channeldropcurves} 
\end{figure*}
\subsection{Comparison with Augmentation-based SDG
Methods}
For comparison with augmentation-based DG methods, we insert MixStyle, EFDMix, TriD, and InstMix (a simple baseline) directly after the raw first layer (pre-BN \texttt{conv1}) to be consistent with our methods' placement, plus the next two full blocks of the ResNet-50 encoder (block2 and block3). FourierAM is applied on the input images directly.

\begin{table*}[!htbp]
\centering
\small
\caption{Comparison of Dice and clDice  for baseline, baseline + WCA, and AngioDG. For baseline + WCA, we experimentally apply a softmax scaling of  $\phi=1e-5$ (other hyperparameters match LA@15 fine-tuning).}
\label{tab:baseline_vs_wca_both_metrics}
\resizebox{\textwidth}{!}{
\begin{tabular}{l cc cc cc cc cc cc}
\toprule
& \multicolumn{2}{c}{SJTU (In-domain)} 
& \multicolumn{2}{c}{OxAMI} 
& \multicolumn{2}{c}{UMAE} 
& \multicolumn{2}{c}{CHUAC} 
& \multicolumn{2}{c}{XCAD} 
& \multicolumn{2}{c}{MTU} \\
\cmidrule(lr){2-3} \cmidrule(lr){4-5} \cmidrule(lr){6-7} \cmidrule(lr){8-9} \cmidrule(lr){10-11} \cmidrule(lr){12-13}
Method 
& Dice & clDice & Dice & clDice & Dice & clDice & Dice & clDice & Dice & clDice & Dice & clDice \\
\midrule
Baseline 
Baseline 
& 83.27 $\pm$ 2.94 & 87.74 $\pm$ 3.09
& 78.87 $\pm$ 7.79 & 82.53 $\pm$ 9.23
& 69.12 $\pm$ 5.82 & 80.93 $\pm$ 6.28
& 69.09 $\pm$ 6.50 & 77.09 $\pm$ 6.53
& 75.34 $\pm$ 5.30 & 78.42 $\pm$ 6.48
& 82.35 $\pm$ 6.71 & 84.11 $\pm$ 7.54 \\
Baseline + WCA 
& 83.69 $\pm$ 2.92 & 88.28 $\pm$ 2.98
& 79.16 $\pm$ 7.68 & 83.06 $\pm$ 9.16
& 68.96 $\pm$ 6.27 & 80.83 $\pm$ 6.60
& 68.84 $\pm$ 6.10 & 77.16 $\pm$ 6.44
& 75.31 $\pm$ 5.46 & 78.43 $\pm$ 6.82
& 82.44 $\pm$ 6.44 & 84.21 $\pm$ 7.15 \\
AngioDG 
& 83.89 $\pm$ 2.97 & 88.36 $\pm$ 3.04
& 79.38 $\pm$ 8.03 & 83.41 $\pm$ 9.12
& 68.90 $\pm$ 6.79 & 80.62 $\pm$ 7.66
& 71.55 $\pm$ 5.29 & 78.78 $\pm$ 5.44
& 75.91 $\pm$ 5.09 & 79.18 $\pm$ 6.41
& 82.76 $\pm$ 7.15 & 84.22 $\pm$ 7.84 \\

\bottomrule
\end{tabular}
}
\end{table*}
In MixStyle and EFDMix, the mixing coefficient \(\lambda\) is sampled  from \(\mathrm{Beta}(\alpha,\alpha)\), where $\alpha=0.1$; similarly, per–channel Bernoulli gating with \(\lambda \sim \mathrm{Beta}(0.1,0.1)\) is used in TriD to swap in \(\mathcal{U}(0,1)\) statistics; for the case of the simple feature-averaging baseline (InstMix) a fixed mixing \(\lambda=0.5\) is used. In terms of FourierAM, we perform Fourier amplitude mixing adapted from the \texttt{colorful\_spectrum\_mix} operator of Xu \textit{et~al.}~\cite{xu2021FACT}, applied directly on input images (added \texttt{@torch.no\_grad()} as it has no preceding layers) with probability $p=0.5$.\footnote{\url{https://github.com/MediaBrain-SJTU/FACT}} We sample a per-pair mixing coefficient $\lambda\sim\mathcal{U}(0,1)$; Fourier magnitudes (ratio $=1.0$, full spectrum) are then mixed, and each image is reconstructed using its original phase. Implementations of MixStyle, EFDMix, and TriD follow TriD’s official repository \footnote{\url{https://github.com/Chen-Ziyang/TriD}}. All augmentation-based DG modules are active during training with probability \(p=0.5\) (using \texttt{torch.rand}), but disabled at test time. We include InstMix  solely as a simple control baseline; no novelty is claimed.

\section{Discussion}

\subsection{Ablation Studies}

As can be observed in Fig.~\ref{channeldropcurves}, the baseline possesses many sharp performance drops for certain channels, which indicates their dominance. Hence, when we fine-tune the baseline directly (without CW), we observe softmax saturation, which can be possibly linked to the combination of concentrated channel importances (highlighting this dominance) and also feature-scale effects.

On the contrary, LA@15 spreads channel importance (we also observe lower feature-scales) and hence for the same fine-tuning settings, i.e. LA@15 with a softmax scaling $\phi=0.1$, LA@15 can be seen to improve with WCA. As for the baseline where some channels dominate and scale effects exist, we need a lower softmax scaling to counteract saturation, where we experimentally find that a softmax scaling of $\phi=1e-5$ (other fine-tuning parameters fixed) gives a good spread and balanced OOD results (we sweep $\phi \in [1e{-3},\,1e{-8}]$). Nevertheless, while it can be argued that this circumvents the issue, even after we use $\phi=1e-5$, baseline + WCA still does not generalize as well for OOD results compared to when we fine-tune on LA@15 (AngioDG), as can be observed in Table~\ref{tab:baseline_vs_wca_both_metrics}.

\section{Experimental Setup}
We ran all experiments on a single NVIDIA GeForce RTX 3060 and fixed the random seed to 42. On our setup,  rerunning our experiments twice for the same configuration produced the same results. 

\clearpage   

{
    \small
    \bibliographystyle{ieeenat_fullname}
    \bibliography{refs}
}